%% file: icml2019.tex
\documentclass[usenames,dvipsnames]{article} 
\pdfoutput=1
\usepackage{times}
\usepackage[accepted,nohyperref]{icml2019}
\usepackage{url}
\usepackage{hyperref}
\usepackage{graphicx}
\usepackage{amsmath,amssymb} 
\usepackage{microtype}
\usepackage{booktabs}
\usepackage{color}
\usepackage[utf8]{inputenc}
\usepackage{graphicx}
\usepackage{verbatim}
\usepackage{subfig}
\usepackage{amsmath}
\usepackage{amsfonts}
\usepackage{mathtools}
\usepackage{multicol}
\usepackage{xspace}
\usepackage{enumitem}
\usepackage[title]{appendix}

\usepackage[table,xcdraw]{xcolor}
\DeclarePairedDelimiterX{\infdivx}[2]{(}{)}{%
  #1\;\delimsize\|\;#2%
}
\usepackage{icml2019}

\newcommand{\infdiv}{\infdivx}

\icmltitlerunning{Learning Disentangled Representations with Reference-Based Variational Autoencoders}

\begin{document}

\def\fig#1{Figure~\ref{fig:#1}}
\def\tab#1{Table~\ref{tab:#1}}
\def\sect#1{Section~\ref{sec:#1}}
\def\Eq#1{Eq.~(\ref{eq:#1})}
\makeatletter
\DeclareRobustCommand\onedot{\futurelet\@let@token\@onedot}
\def\@onedot{\ifx\@let@token.\else.\null\fi\xspace}
\def\eg{\emph{e.g}\onedot} \def\Eg{\emph{E.g}\onedot}
\def\ie{\emph{i.e}\onedot} \def\Ie{\emph{I.e}\onedot}
\def\cf{\emph{c.f}\onedot} \def\Cf{\emph{C.f}\onedot}
\def\etc{\emph{etc}\onedot} \def\vs{\emph{vs}\onedot} 
\def\wrt{w.r.t\onedot} \def\dof{d.o.f\onedot}
\def\etal{\emph{et al}\onedot}

\twocolumn[
\icmltitle{Learning Disentangled Representations with \\
Reference-Based Variational Autoencoders}
\icmlsetsymbol{equal}{*}

\begin{icmlauthorlist}
\icmlauthor{Adria Ruiz}{inria}
\icmlauthor{Oriol Martinez}{upf}
\icmlauthor{Xavier Binefa}{upf}
\icmlauthor{Jakob Verbeek}{inria}
\end{icmlauthorlist}

\icmlaffiliation{inria}{Univ. Grenoble Alpes, Inria, CNRS, Grenoble INP, LJK, 38000 Grenoble, France}
\icmlaffiliation{upf}{Univ. Pompeu Fabra, DTIC , 08018 Barcelona, Spain}

\icmlcorrespondingauthor{Adria Ruiz}{adria.ruiz-ovejero@inria.fr}

\icmlkeywords{Disentangled representations, Variational Autoencoders, Adversarial Learning, Weakly-supervised learning}

\vskip 0.3in
]
\printAffiliationsAndNotice{} 
\newcommand{\fix}{\marginpar{FIX}}
\newcommand{\new}{\marginpar{NEW}}

\begin{abstract}
Learning disentangled representations from visual data, where different high-level generative factors are independently encoded, is of importance for many computer vision tasks. Solving this problem, however, typically requires to explicitly label all the  factors of interest in training images. To alleviate the annotation cost, we introduce a learning setting which we refer to as \textit{reference-based disentangling}. Given a pool of unlabelled images, the goal is to learn a representation where a set of target factors are disentangled from others. The only supervision comes from an auxiliary \textit{reference set} containing  images where the factors of interest are constant. In order to address this problem, we propose reference-based variational autoencoders, a novel deep generative model designed to exploit the weak-supervision provided by the reference set. By addressing tasks such as feature learning, conditional image generation or attribute transfer, we validate the ability of the proposed model to learn disentangled representations from this minimal form of supervision.
\end{abstract}
\input{introduction.tex}

\input{related.tex}

\input{method.tex}

\input{experiments.tex}
\section{Conclusions}
In this paper we have introduced the reference-based disentangling problem and proposed reference-based variational autoencoders to address it. We have shown that the standard variational learning objective used to train VAE can lead to degenerate solutions  when it is applied in our setting, and proposed an alternative training strategy that exploits adversarial learning. Comparing the proposed model with previous state-of-the-art approaches, we have shown its ability to learn disentangled representations from minimal supervision and its application to tasks such as feature learning, conditional image generation and attribute transfer.



\bibliographystyle{icml2019}
\bibliography{biblio,jjv}

\clearpage
\begin{appendices}
\onecolumn
\input{supplementary.tex}

\end{appendices}

\end{document}

%% file: introduction.tex
\section{Introduction}
Natural images are the result of a generative process involving a large number factors of variation. For instance, the appearance of a face is determined by the interaction between many latent variables including the pose, the illumination, identity, and  expression.  Given that the interaction between these underlying explanatory factors is very complex, inverting the generative process is extremely challenging.

From this perspective, learning disentangled representations where different high-level generative factors are independently encoded can be considered one of the most relevant problems in computer vision \citep{bengio2013representation}. For instance, these representations can be applied to complex classification tasks given that features correlated with image labels can be easily identified. We find another example in conditional image generation \citep{van2016conditional,yan2016attribute2image}, where disentangled representations allow to manipulate high-level attributes in synthesized images. 

\textbf{Motivation:} By coupling deep learning with variational inference, Variational autoencoders  (VAEs)~\citep{kingma2013auto} have emerged as a powerful latent variable model able to learn abstract data representations. However, VAEs are typically trained in an unsupervised manner and, they therefore lack a mechanism to impose specific high-level semantics on the latent space. In order to address this limitation, different semi-supervised variants have been  proposed \citep{kingma2014semi,narayanaswamy2017learning}.  These approaches, however, require latent factors to be explicitly labelled in a training set. These annotations provide supervision to the model, and allow to disentangle the labelled variables from the remaining generative factors.  The main drawback of this strategy is that it may require a significant annotation effort. For instance, if we are interested in disentangling  facial gesture information from face images, we need to annotate samples according to different expression classes. While this is feasible for a reduced number of basic gestures, natural expressions depend on a combination of a large number of facial muscle activations with their corresponding intensities \citep{ekman1997face}. Therefore, it is impractical to label all these factors even in a small subset of training images. In this context, our main motivation is to explore a novel learning setting allowing to disentangle specific factors of variation while minimizing the required annotation effort. 

\begin{figure}[t]
\centering
\includegraphics[width=\linewidth]{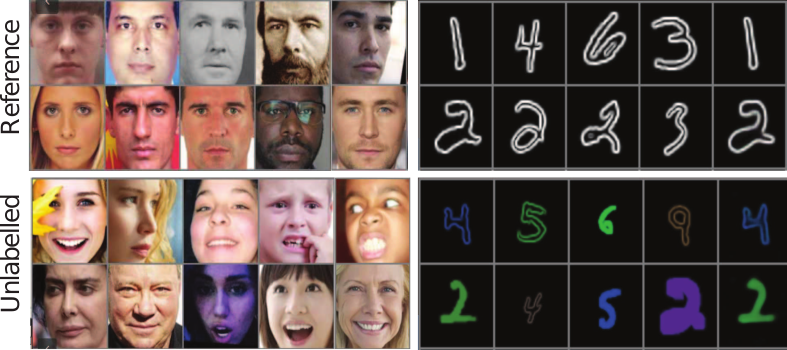}
\caption{Examples of reference-based disentangling problems. 
{Left:~Disentangling} factors underlying  facial expression. The reference set contains faces with neutral expression.
Right: Disentangling style from digits. The reference set is composed by digits with a fixed style.
}
\label{fig:rbd_example}
\end{figure}

\textbf{Contributions:} We introduce \textit{reference-based disentangling}. A learning setting in which, given a training set of unlabelled images, the goal is to learn a  representation where a specific set of generative factors are disentangled from the rest. For that purpose, the only supervision comes in the form of an auxiliary \textit{reference set} containing images where the factors of interest are constant (see Fig. \ref{fig:rbd_example}). Different from a semi-supervised scenario, explicit labels are not available for the factors of interest during training. In contrast, reference-based disentangling is a weakly-supervised task, where the reference set only provides implicit information about the generative factors that we aim to disentangle. Note that a collection of reference images is generally easier to obtain compared to explicit labels of target factors. For example, it is more feasible to collect a set of faces with a neutral expression, than to annotate images across a large range of expression classes or attributes. 

The main contributions of our paper are summarized as follows: \textbf{(1)} We propose reference-based variational autoencoders (Rb-VAEs). Different from unsupervised VAEs, our model is able to impose high-level semantics into the latent variables by exploiting the weak supervision provided by the reference set; \textbf{(2)} We identify critical limitations of the standard VAE objective when used  to train our  model. To address this problem, we propose an alternative training procedure based on recently introduced ideas in the context of variational inference and adversarial learning; \textbf{(3)} By learning disentangled representations from minimal supervision, we show how our framework is able to naturally address tasks such as feature learning, conditional image generation, and attribute transfer.


%% file: related.tex
\section{Related Work}
\label{sec:related}
\textbf{Deep Generative Models} have been extensively explored to model visual and other types of data. Variational autoencoders~\citep{kingma2013auto} and generative adversarial networks (GANs) \citep{goodfellow2014generative} have emerged as two of the most effective frameworks. 
VAEs use variational evidence lower bound  to learn an encoder network that maps images to an approximation of the  posterior distribution over latent variables. Similarly, a decoder network is learned that produces  the conditional distribution on images given the  latent variables. GANs are also composed of two differentiable networks. The generator network synthesizes images from latent variables, similar to the VAE decoder. The discriminator's goal is to separate  real training images from  synthetic images sampled from the generator. During training, GANs employ an adversarial learning procedure which allows to simultaneously optimize the discriminator and generator parameters. Even though GANs have been shown to generate more realistic samples than VAEs, they lack an inference mechanism able to map images into their corresponding latent variables. In order to address this drawback, there have been several attempts to combine ideas from VAEs and GANs~\citep{larsen2015autoencoding,dumoulin2016adversarially,donahue2016adversarial}. Interestingly, it has been shown that adversarial learning can be used to minimize the variational objective function of VAEs~\citep{makhzani2015adversarial,huszar2017variational}. Inspired by this observation, various methods such as adversarial variational Bayes~\citep{mescheder2017adversarial}, $\alpha$-GAN \citep{rosca2017variational}, and symmetric-VAE (sVAE) \citep{pu2017symmetric} have incorporated adversarial learning into the VAE framework.

Different from this prior work, our Rb-VAE model is a deep generative model specifically designed to solve the reference-based disentangling problem. During training, adversarial learning is used in order to minimize a variational objective function inspired by the one employed in sVAE~\citep{pu2017symmetric}.  Although sVAE  was originally motivated by the limitations of the maximum likelihood criterion used in unsupervised VAEs, we show how its variational formulation offers specific advantages in our context.

\textbf{Learning Disentangled Representations} is a long standing problem in machine learning and computer vision~\citep{bengio2013representation}. In the literature, we can differentiate three main paradigms  to address it: unsupervised, supervised, and weakly-supervised learning. Unsupervised models are trained without specific information about the generative factors of interest~\citep{desjardins2012disentangling,chen2016infogan}. To address this task, the most common approach consists in imposing different constraints on the latent representation. For instance, unsupervised VAEs typically define the prior over the latent variables with a fully-factorized Gaussian distribution. Given that high-level generative factors are typically independent, this prior encourage their disentanglement in different dimensions of the latent representation. Based on this observation, different approaches such as $\beta$-VAE \citep{higgins2016beta}, DIP-VAE \citep{kumar2017variational}, FactorVAE \cite{kim2018disentangling} or $\beta$-TCVAE \citep{chen2018isolating} have explored more sophisticated regularization mechanisms over the distribution of inferred latent variables. Although unsupervised approaches are able to identify simple explanatory components, they do not allow latent variables to model specific high-level factors. 

A straight-forward approach to overcome this limitation is to use a fully-supervised strategy. In this scenario, models are learned by using a training set where the factors of interest are explicitly labelled. Following this paradigm, we can find different semi-supervised \citep{kingma2014semi,narayanaswamy2017learning}, and conditional \citep{yan2016attribute2image,pu2016variational} variants of autoencoders. In spite of the effectiveness of supervised approaches in different applications, obtaining explicit labels is not feasible in scenarios where we aim to disentangle a large number of factors or their annotation is difficult. An intermediate solution between unsupervised and fully-supervised methods are weakly-supervised approaches. In this case, only implicit information about factors of variation is provided during training. Several works have explored this strategy by using different forms of weak-supervision such as: temporal coherence in sequential data \citep{hsu2017unsupervised,denton2017unsupervised,villegas2017decomposing}, pairs of aligned images obtained from different domains \cite{gonzalez2018image} or knowledge about the rendering process in computer graphics~\citep{yang2015weakly,kulkarni2015deep}.

Different from previous works relying on other forms of weak supervision, our method addresses the reference-based disentangling problem. In this scenario, the challenge is to exploit the implicit information provided by a training set of images where the generative factors of interest are constant. Related with this setting, recent approaches have considered to exploit pairing information of images known to  share the same generative factors \citep{mathieu2016disentangling,donahue2017semantically,feng2018dual,bouchacourt2017multi}. However, the amount of supervision required by these methods is larger than the available in reference-based disentangling. Concretely, we only know that reference images are generated by the same constant factor. In addition, no information is available about what unlabelled samples share the same target factors. 


%% file: method.tex
\section{Preliminaries: Variational Autoencoders}
\label{sec:vaes}

Variational autoencoders (VAEs) are  generative models defining a joint distribution $p_\theta(\mathbf{x},\mathbf{z}) = p_\theta(\mathbf{x}|\mathbf{z})p(\mathbf{z})$, where $\mathbf{x}$ is an observation, \eg an image, and $\mathbf{z}$ is a latent variable with a simple prior $p(\mathbf{z})$, \eg a Gaussian  with zero mean and identity covariance matrix. 
Moreover, $p_\theta(\mathbf{x}|\mathbf{z})$ is typically modeled as a factored Gaussian, whose mean and diagonal covariance matrix are given by a function of $\mathbf{z}$, implemented by a \textit{generator} neural network.

Given a training set of samples from an unknown data distribution $p(\mathbf{x})$, VAEs learn the optimal parameters $\theta$ by defining a variational distribution $q_\psi(\mathbf{x},\mathbf{z})=q_\psi(\mathbf{z}|\mathbf{x}) p(\mathbf{x})$. Note that  $q_\psi(\mathbf{z}|\mathbf{x})$ approximates the intractable posterior $p(\mathbf{z} | \mathbf{x})$ and  is defined as another factored Gaussian, whose mean and diagonal covariance matrix are given as the output of an \textit{encoder} or \textit{inference} network with parameters $\psi$. 
The generator and the encoder are optimized by solving:
\begin{equation}
    \min_{\theta, \psi}  \mathbb{E}_{p(\mathbf{x})}\big[ \mathbb{KL}\infdiv{q_\psi(\mathbf{z}|\mathbf{x})}{p(\mathbf{z})}  - \mathbb{E}_{q_\psi(\mathbf{z}|\mathbf{x})}\log(p_\theta(\textbf{x}|\mathbf{z})) \big], \nonumber
    \label{eq:VAEMin}
\end{equation}
which is equivalent to the minimization of the KL divergence between $q_\psi(\mathbf{x},\mathbf{z})$ and $p_\theta(\mathbf{x},\mathbf{z})$. The first KL term can be interpreted as a regularization mechanism encouraging the distribution $q_\psi(\mathbf{z}|\mathbf{x})$ to be similar to $p(\mathbf{z})$. The second term is known as the reconstruction error, measuring the negative log-likelihood of a generated sample $\mathbf{x}$ from its latent variables $q_\psi(\mathbf{z}|\mathbf{x})$. Optimization can be carried out by using stochastic gradient descent (SGD) where $p(\mathbf{x})$ is approximated by the training set. The \textit{re-parametrization trick} ~\citep{rezende14icml} is employed to enable gradient back-propagation across samples from $q_\phi(\mathbf{z}|\mathbf{x})$.

\section{Reference-based Disentangling}
\label{sec:Rb-VAEs}

Consider a training set of unlabelled images (\eg human faces) $\mathbf{x} \in \mathbb{R}^{W \times H \times 3}$  sampled from a given distribution $p^u(\mathbf{x})$. Our goal is to learn a latent variable model defining a joint distribution over $\mathbf{x}$ and latent variables $\mathbf{e} \in \mathbb{R}^{D_e}$ and $\mathbf{z} \in \mathbb{R}^{D_z}$. Whereas $\mathbf{e}$ is expected to encode information about a set of generative factors of interest, \eg facial expressions, $\mathbf{z}$ should model the remaining factors of variation underlying the images, \eg pose, illumination, identity, \etc. From now on, we will refer to $\mathbf{e}$ and $\mathbf{z}$ as the ``target'' and ``common factors'', respectively. In order to disentangle them, we are provided with an additional set of reference images sampled from $p^r(\mathbf{x})$, representing a distribution over $\mathbf{x}$ where target factors $\mathbf{e}$ are constant \eg neutral faces. Given $p^r(\mathbf{x})$ and $p^u(\mathbf{x})$, we define an auxiliary binary variable $y \in \{0,1\}$ indicating whether an image $\mathbf{x}$ has been sampled from the unlabelled or reference distributions, \ie  $p(\mathbf{x}|y=0)=p^u(\mathbf{x})$ and $p(\mathbf{x}|y=1)=p^r(\mathbf{x})$. In reference-based disentangling, we aim to exploit the weak-supervision provided by  $y$ in order to effectively disentangle target factors $\mathbf{e}$ and common factors $\mathbf{z}$.

\subsection{Reference-based Variational Autoencoders}

\begin{figure}[t]
\centering
\includegraphics[width=0.9\linewidth]{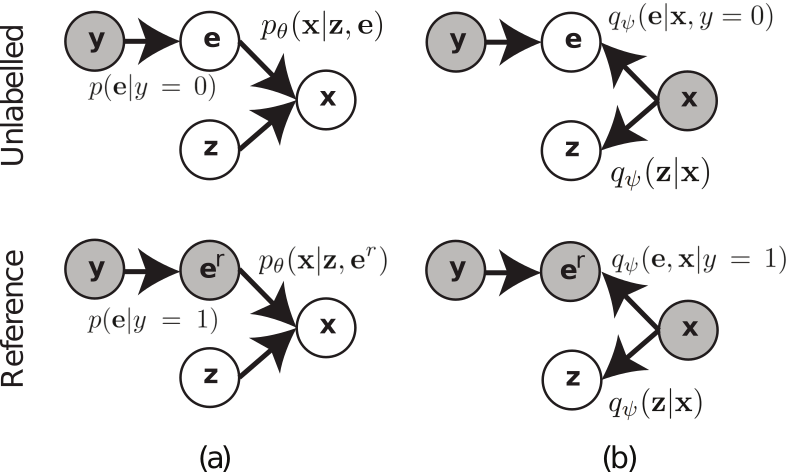}
\caption{(a) Rb-VAE generative process where $p_\theta(\mathbf{x}|\mathbf{z},\mathbf{e})$ maps latent variables $\mathbf{z}$ (common factors) and $\mathbf{e}$ (target factors) to images $\mathbf{x}$. Shaded circles indicate observed variables. Note that for the reference samples $(y=0)$, the prior $p(\mathbf{e}|y)$ is deterministic given that images are
are known to be generated by constant $\mathbf{e}^r$. (b) Approximate posteriors  $q(\mathbf{z}|x)$ and $q(\mathbf{e}|\mathbf{x},y)$ map images $\mathbf{x}$ to the corresponding common and target factors $\mathbf{z}$ and $\mathbf{e}$ respectively.} 
\label{fig:overview}
\end{figure}

In this section, we present reference-based variational auto-encoders (Rb-VAE). Rb-VAE is a  deep latent variable model defining a joint distribution:
\begin{equation}
p_\theta(\mathbf{x},\mathbf{z},\mathbf{e},y) = p_\theta(\mathbf{x} | \mathbf{z} , \mathbf{e}) p(\mathbf{z}) p(\mathbf{e}|y)p(y),
\end{equation}
where conditional dependencies are designed to address the reference-based disentangling problem, see Fig.~\ref{fig:overview}(a). We define $p_\theta(\mathbf{x} | \mathbf{z} , \mathbf{e})=\mathcal{L}(\mathbf{x} | \mathcal{G}_\theta(\mathbf{z}, \mathbf{e}), \lambda )$, where $\mathcal{G}_\theta(\mathbf{z}, \mathbf{e})$ is the generator network, mapping a pair of latent variables $(\mathbf{z,e})$ to an image defining the mean of a Laplace distribution  $\mathcal{L}$ with fixed scale parameter $\lambda$. 
We use a Laplace distribution, instead of the Gaussian usually employed in the VAEs. The reason is that the negative log-likelihood is equivalent to the $\ell_1$-loss which encourages sharper image reconstructions with better visual quality \citep{mathieu2015deep}.

To reflect the assumption of constant  target factors  across reference images, we 
define the conditional distribution over $\mathbf{e}$ given  $y=1$ as a delta peak centered on a learned vector $\mathbf{e}^r \in R^{D_e}$, \ie  $p(\mathbf{e}|y=1)=\delta(\mathbf{e}-\mathbf{e}^r)$. In contrast, for $y=0$, the conditional distribution is set to a unit Gaussian, $p(\mathbf{e}|y=0)=\mathcal{N}(\mathbf{e} | \mathbf{0},\mathbf{I})$, as in standard VAEs. In the following, we denote $p(\mathbf{e} | y=0)=p(\mathbf{e})$.  Contrary to the case of target factors $\mathbf{e}$, the prior over common factors $\mathbf{z}$ is equal for reference and unlabelled images, and taken to be a unit Gaussian $p(\mathbf{z})= \mathcal{N}(\mathbf{z} | \mathbf{0},\mathbf{I})$. Finally, we assume a uniform prior over $y$, \ie $p(y=0)=p(y=1)=\frac{1}{2}$.

\subsection{Conventional Variational Learning}
\label{sec:rbvae-opt}
Following the standard VAE framework discussed in Sec.~\ref{sec:vaes}, we define a variational distribution $q_\psi(\mathbf{x}, \mathbf{z},\mathbf{e},y)= q_\psi(\mathbf{x}, \mathbf{z} | \mathbf{x}) q_\psi( \mathbf{e} | \mathbf{x} , y) p(\mathbf{x},y)$, and learn the model parameters $\theta$ by minimizing the KL divergence between $q_\psi$ and $p_\theta$:
\begin{equation}
\min_{\theta, \psi} \hspace{3mm} \mathbb{KL}\infdiv{q_\psi(\mathbf{x},\mathbf{z},\mathbf{e},y)}{p_\theta(\mathbf{x},\mathbf{z},\mathbf{e},y)}.    
\label{eq:rbvae_kl}
\end{equation}
Note that the conditionals $q_\psi(\mathbf{e} | \mathbf{x}, y)$ and $q_\psi(\mathbf{z} | \mathbf{x})$ provide a factored approximation of  the intractable posterior $p_\theta(\mathbf{e,z} | \mathbf{x},y)$, allowing to infer target and common factors $\mathbf{e}$ and $\mathbf{z}$ given the image $\mathbf{x}$, see Fig.~\ref{fig:overview}(b). 
Given a reference image, \ie with $y=1$, the target factors $q_\psi(\mathbf{e} | \mathbf{{x}}, y=1)$ are known to be equal to the reference value $\mathbf{e}^r$. On the other hand, given an non-reference image, \ie with $y=0$, we define the approximate posterior $q_\psi(\mathbf{e} | \mathbf{{x}}, y=0) = \mathcal{N}(\mathbf{e} | \mathcal{E}^\mu(\mathbf{{x}}), \mathcal{E}^\sigma(\mathbf{{x}}))$, where the means and diagonal covariance matrices of a conditional Gaussian distribution are given by non-linear functions  $\mathcal{E}^\mu(\mathbf{{x}})$ and $\mathcal{E}^\sigma(\mathbf{{x}})$, respectively. Similarly, we use an additional network to model $q_\psi(\mathbf{z} | \mathbf{{x}}) = \mathcal{N}(\mathbf{z} | \mathcal{Z}^\mu(\mathbf{{x}}), \mathcal{Z}^\sigma(\mathbf{{x}}))$.

\textbf{Optimization.} In Appendix \ref{ap:maths} we show  that the minimization of Eq.~(\ref{eq:rbvae_kl}) can be expressed as
\begin{align}
\min_{\theta, \psi, \mathbf{e}^r} \hspace{3mm} & \mathbb{E}_{ p^u(\mathbf{x})} \Big[ \mathbb{KL}\infdiv{q_\psi( \mathbf{z} | \mathbf{x}) q_\psi( \mathbf{e} | \mathbf{x})}{p(\mathbf{z}) p(\mathbf{e})} - \nonumber \\ 
& \hspace{11mm} \mathbb{E}_{ q_\psi( \mathbf{z} | \mathbf{x}) q_\psi( \mathbf{e} | \mathbf{x})}  \log(p_\theta(\mathbf{x} | \mathbf{z} , \mathbf{e}))\Big] + \nonumber \\
&  \mathbb{E}_{ p^r(\mathbf{x})} \Big[ \mathbb{KL}\infdiv{q_\psi( \mathbf{z} | \mathbf{x})}{p(\mathbf{z})} - \nonumber \\ 
& \hspace{11mm} \mathbb{E}_{ q_\psi( \mathbf{z} | \mathbf{x})}  \log(p_\theta(\mathbf{x} | \mathbf{z} , \mathbf{e}^r))\Big],
\label{eq:RBVAEKLMin_Optimization}
\end{align}
where the second and fourth terms of the expression correspond to the reconstruction errors for unlabelled and reference images respectively. Note that, for reference images, no inference over target factors $\mathbf{e}$ is needed. Instead, the generator reconstructs them  using the learned parameter $\mathbf{e}^r$. Similar to standard VAEs, the remaining terms consist of KL divergences between approximate posteriors and priors over the latent variables. The minimization problem defined in Eq.~(\ref{eq:RBVAEKLMin_Optimization}) can be solved using SGD and the re-parametrization trick in order to back-propagate the gradient when sampling from $q_\psi(\mathbf{e} | \mathbf{x})$ and $q_\psi(\mathbf{z} | \mathbf{x})$.

\subsection{Symmetric Variational Learning}
\label{sec:symmetric_learning}

The main limitation of the variational objective defined in \Eq{RBVAEKLMin_Optimization} is that it does not guarantee that common and target factors will be effectively disentangled in $\mathbf{z}$ and $\mathbf{e}$ respectively. In order to understand this phenomenon, it is necessary to analyze the role of the conditional distribution $p(\mathbf{e}|y)$ in Rb-VAEs. By defining $p(\mathbf{e}|y=1)$ as a delta function, the model is forced to encode into $\mathbf{z}$ all the generative factors of reference images, given that they must be reconstructed via $p_\theta(\mathbf{x} | \mathbf{z}, \mathbf{e}^r)$ with constant $\mathbf{e}^r$. Therefore, $p(\mathbf{e}|y)$ is implicitly encouraging $q_\psi(\mathbf{z} | \mathbf{x})$ to encode common factors present in reference and unlabelled samples. However, this mechanism does not avoid the scenario where target factors are also encoded into latent variables $\mathbf{z}$. More formally, given that $ \mathbf{z}$ is expressive enough, the minimization of Eq.~(\ref{eq:RBVAEKLMin_Optimization}) does not prevent a degenerate solution $p_\theta(\mathbf{x} | \mathbf{z}, \mathbf{e})=p_\theta(\mathbf{x} | \mathbf{z})$, where the inferred latent variables by $q_\psi(\mathbf{e} | \mathbf{x})$ are ignored by the generator. 

To address this  limitation, we propose to optimize an alternative variational expression inspired by unsupervised Symmetric VAEs~\citep{pu2017symmetric}. Specifically, we add the reverse KL between $q_\psi$ and $p_\theta$ to the objective of the minimization problem:
\begin{align}
\min_{\theta, \psi} \hspace{3mm} \mathbb{KL}\infdiv{q_\psi(\mathbf{x},\mathbf{z},\mathbf{e},y)}{p_\theta(\mathbf{x},\mathbf{z},\mathbf{e},y)} + & \nonumber \\ \mathbb{KL}\infdiv{p_\theta(\mathbf{x},\mathbf{z},\mathbf{e},y)}{q_\psi(\mathbf{x},\mathbf{z},\mathbf{e},y)}. 
\label{eq:SRBVAEKLMin}
\end{align}
In order to understand why this additional term allows to mitigate the degenerate solution $p_\theta(\mathbf{x} | \mathbf{z},\mathbf{e}) = p_\theta(\mathbf{x} | \mathbf{z})$, it is necessary to observe that its minimization is equivalent to:
\begin{align}
\label{eq:SRBVAEKLMin_Optimization}
\min_{\theta, \psi} \hspace{2mm} & 
\mathbb{E}_{ p(\mathbf{z},\mathbf{e})} \Big[ \mathbb{KL}\infdiv{p_\theta(\mathbf{x} | \mathbf{z} , \mathbf{e})}{p^u(\mathbf{x})} - \\
& \hspace{12mm} \mathbb{E}_{p_\theta(\mathbf{x} | \mathbf{z} , \mathbf{e})} [\log(q_\psi( \mathbf{z} | \mathbf{x})) + \log(q_\psi( \mathbf{e} | \mathbf{x}))] \Big] + \nonumber \\
& \mathbb{E}_{p(\mathbf{z}) p_\theta(\mathbf{x} | \mathbf{z} , \mathbf{e}^r)} \Big[ \mathbb{KL}(p_\theta(\mathbf{x} | \mathbf{z} , \mathbf{e}^r) || p^r(\mathbf{x})) - \log(q_\psi( \mathbf{z} | \mathbf{x})) \Big], \nonumber
\end{align}
see  Appendix~\ref{ap:maths} for details. Note that the two KL divergences encourage images generated using $p(\mathbf{z})$, $p(\mathbf{e})$ and $\mathbf{e}^r$ to be similar to samples from the real distributions $p^r(\mathbf{x})$ and $p^u(\mathbf{x})$. On the other hand, the remaining terms correspond to reconstruction errors over latent variables ${\mathbf{z},\mathbf{e}\\}$ inferred from generated images drawn from $p_\theta$. As a consequence, the minimization of these errors is encouraging the generator $p_\theta(\mathbf{x} | \mathbf{z},\mathbf{e})$ to generate images $\mathbf{x}$ by taking into account latent variables $\mathbf{e}$, since the latter   must be reconstructed via $q_\psi(\mathbf{e} | \mathbf{x})$. In conclusion, the minimization of the reversed KL avoids the degenerate solution ignoring  $\mathbf{e}$.

\textbf{Optimization via Adversarial Learning.}
\label{sec:opt_adv_learning}
Given the introduction of the reversed KL divergence, the learning procedure described in Sec. \ref{sec:rbvae-opt} can not be directly applied to the minimization of \Eq{SRBVAEKLMin}. However, note that we can express the defined symmetric objective as:
\begin{align}
\min_{\theta,\psi} \hspace{3mm} & \mathbb{E}_{q_\psi(\mathbf{e} , \mathbf{z} | \mathbf{x}) p^u(\mathbf{x})} \mathcal{L}_{\mathbf{x}\mathbf{z}\mathbf{e}} -  \mathbb{E}_{p_\theta(\mathbf{x}|\mathbf{e},\mathbf{z}) p(\mathbf{z}) p(\mathbf{e}) } \mathcal{L}_{\mathbf{x}\mathbf{z}\mathbf{e}} \nonumber \\
& + \mathbb{E}_{q_\psi(\mathbf{z} | \mathbf{x}) p^r(\mathbf{x}) } \mathcal{L}_{\mathbf{x}\mathbf{z}} - \mathbb{E}_{p_\theta(\mathbf{x}|\mathbf{e}^r,\mathbf{z}) p(\mathbf{z})} \mathcal{L}_{\mathbf{x}\mathbf{z}},
\label{eq:srbvae_optimization}
\end{align}
where $\mathcal{L}_{\mathbf{x}\mathbf{z}\mathbf{e}}$ corresponds to the log-density ratio between distributions $q_\psi(\mathbf{e} , \mathbf{z} | \mathbf{x}) p^u(\mathbf{x})$ and $p_\theta(\mathbf{x}|\mathbf{e},\mathbf{z}) p(\mathbf{z}) p(\mathbf{e})$. Similarly, $\mathcal{L}_{\mathbf{x}\mathbf{z}}$ defines an analogous expression for $q_\psi(\mathbf{z} | \mathbf{x}) p^r(\mathbf{x})$ and $p_\theta(\mathbf{x}|\mathbf{e}^r,\mathbf{z}) p(\mathbf{z})$. See Appendix \ref{ap:maths} for a detailed derivation.

Taking into account previous definitions, SGD optimization can be employed in order to learn model parameters. Concretely, we can evaluate $\mathcal{L}_{\mathbf{x}\mathbf{z}\mathbf{e}}$ and $\mathcal{L}_{\mathbf{x}\mathbf{z}}$ to back-propagate the gradients \wrt parameters $\psi$ and $\theta$ by using the re-parametrization trick over samples of $\mathbf{x}$, $\mathbf{e}$ and $\mathbf{z}$. The main challenge of this strategy is that expressions $\mathcal{L}_{\mathbf{x}\mathbf{z}\mathbf{e}}$ and $\mathcal{L}_{\mathbf{x}\mathbf{z}}$ can not be explicitly computed. 
However, the log-density ratio between two distributions can be estimated by using logistic regression \citep{bickel2007discriminative}. In particular, we define an auxiliary parametric function $d_\xi(\mathbf{x},\mathbf{z},\mathbf{e}) \sim \mathcal{L}_{\mathbf{x}\mathbf{z}\mathbf{e}}$ and learn its parameters $\xi$ by solving:
\begin{align}
\max_{\xi} \hspace{3mm} &  \mathbb{E}_{p_\theta(\mathbf{x} | \mathbf{z} , \mathbf{e}) p(\mathbf{z},\mathbf{e})} \log(\sigma(d_\xi(\mathbf{x},\mathbf{z},\mathbf{e})) \nonumber \\ 
& + \mathbb{E}_{q_\psi(\mathbf{e}, \mathbf{z} | \mathbf{x} ) p^u(\mathbf{x})} \log(1-\sigma(d_\xi(\mathbf{x},\mathbf{z},\mathbf{e})), 
\label{eq:disc_optimization}
\end{align}
where $\sigma(\cdot)$ refers to the sigmoid function. Similarly,  $\mathcal{L}_{\mathbf{x}\mathbf{z}}$ is approximated with an additional function $d_\gamma(\mathbf{x},\mathbf{z})$. 

\begin{figure}[t]
\centering
\includegraphics[width=\linewidth]{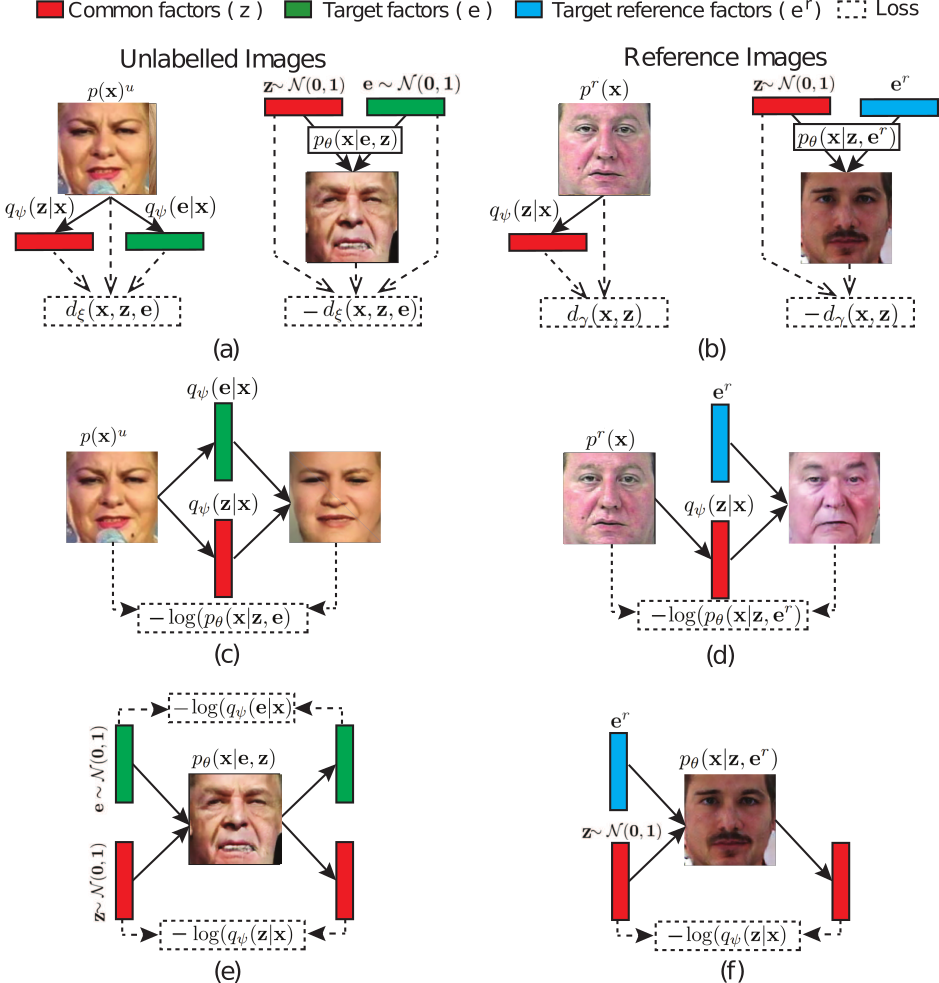}
\caption{Losses used by sRB-VAE.
Discriminator $d_\xi(\mathbf{x},\mathbf{z},\mathbf{e})$ measures the log-density ratio between the  distributions $q_\psi(\mathbf{z},\mathbf{e} | \mathbf{x}) p^u(\mathbf{x})$ and $p_\theta(\mathbf{x}|\mathbf{e},\mathbf{z})p(\mathbf{z})p(\mathbf{e})$.  
(b) Similar loss for reference images using an additional discriminator $d_\gamma(\mathbf{x},\mathbf{z})$ 
(c,d) Reconstruction errors for unlabelled and reference images. (e,f) Reconstruction error over latent variables inferred from unlabelled and reference images generated using $p(\mathbf{z})$, $p(\mathbf{e})$ and $\mathbf{e^r}$}
\label{fig:train_summary}
\end{figure}

This approach is analogous to adversarial unsupervised methods such as ALI \citep{dumoulin2016adversarially}, where the function $d_\gamma(\cdot)$ acts as a discriminator trying to distinguish whether pairs of reference images $\mathbf{x}$ and latent variables $\mathbf{z}$ have been generated by $q_\psi$ and $p_\theta$. However, in our case we have an additional discriminator $d_\xi$ operating over unlabelled images and its corresponding latent variables $\mathbf{z}$ and $\mathbf{e}$ (see Fig. \ref{fig:train_summary}a-b) 
To conclude, it is also interesting to observe that the discriminator $d_\gamma(\mathbf{x},\mathbf{z})$ is implicitly encouraging latent variables $\mathbf{z}$ to encode only information about the common factors. The reason is that samples generated from $p_\theta(\mathbf{x}|\mathbf{z},\mathbf{e^r})p(\mathbf{z})$ are forced to be similar to reference images. As a consequence, $\mathbf{z}$ can not contain information about target factors, which must be encoded into $\mathbf{e}$.

Using previous definitions, we use an adversarial procedure where model and discriminators parameters ($\theta$,$\psi$), and ($\xi$,$\gamma$) are simultaneously optimized by minimizing and maximizing equations (\ref{eq:srbvae_optimization}) and  (\ref{eq:disc_optimization}) respectively. The algorithm used to process one batch during SGD is shown in Appendix~\ref{ap:alg}. In Rb-VAEs, the discriminators $d_\gamma(\cdot)$ and  $d_\xi(\cdot)$ are also implemented as deep convolutional networks.  

\textbf{Explicit Log-likelihood Maximization.}
As shown in equations (\ref{eq:RBVAEKLMin_Optimization}) and (\ref{eq:SRBVAEKLMin_Optimization}), the minimization of the symmetric KL divergence encourages low reconstruction errors for images and inferred latent variables. However, by using the proposed adversarial learning procedure, the minimization of these terms becomes implicit. As shown by \citet{dumoulin2016adversarially} and \citet{donahue2016adversarial}, this can cause original samples to differ substantially from their corresponding reconstructions. In order to address this drawback, we use a similar strategy as \citet{pu2017symmetric} and \citet{li2017alice}, and explicitly add the reconstruction terms into the learning objective, minimizing them together with Eq.~(\ref{eq:srbvae_optimization}), see Fig.~\ref{fig:train_summary}(c--f). In preliminary experiments, we found that the explicit addition of these reconstructions terms during training is important to achieve low reconstruction errors, and to increase stability of adversarial training.


%% file: experiments.tex
\section{Experiments}

\subsection{Datasets}
\label{sec:datasets} 
To validate our approach and to compare to existing work, we consider two different problems.

\textbf{Digit Style Disentangling.} The goal is to model style variations from hand-written digits. We consider the digit style as a set of three different properties: scale, width and color. In order to address this task from a reference-based perspective, we use half of the original  training images in the MNIST  dataset \citep{lecun1998gradient} as our reference distribution (30k examples). The unlabelled set is synthetically generated by applying different transformations over the remaining half of images: (1) Simulation of stroke widths by using a dilation with a given filter size; (2) Digit colorization by multiplying the RGB components of the pixels in an image by a random 3D vector; (3) Size variations by down-scaling the image by a given factor. We randomly transform each image twice to obtain a total of 60k unsuperivsed images. See more details in Appendix \ref{ap:datasets}. 

\textbf{Facial Expression Disentangling.} 
We address the disentangling of facial expressions by using a reference set of neutral faces.
As unlabelled images we use a subset of the AffectNet dataset~\citep{mollahosseini2017affectnet}, which contains a large quantity of facial images. This database is especially challenging since faces were collected ``in the wild'' and exhibit a large variety  of natural expressions. A subset of the images are annotated according to different facial expressions: \emph{happiness, sadness, surprise, fear, disgust, anger}, and \emph{contempt}. We use these labels only for quantitative evaluation. Given that we found that many neutral images in the original database were not correctly annotated, we collected a separate reference set, see Appendix~\ref{ap:datasets}. The unlabelled and reference sets consist of 150k and 10k images, respectively. 

\subsection{Baselines and Implementation Details} 

We evaluate the two different variants of our proposed method: Rb-VAE, trained using the standard variational objective  (Sec. \ref{sec:rbvae-opt}), and sRb-VAE, learned by minimizing the symmetric KL divergence   (Sec. \ref{sec:symmetric_learning}). To demonstrate the advantages of exploiting the weak-supervision provided by reference images, we compare both methods with various state-of-the-art unsupervised approaches based on the VAE framework: $\beta$-VAE~\citep{higgins2016beta}, $\beta$-TCVAE ~\citep{chen2018isolating},  sVAE~\citep{pu2017symmetric}, DIP-VAE-I and DIP-VAE-II ~\citep{kumar2017variational}. Note that $\beta$-VAE DIP-VAE and $\beta$-TCVAE have been specifically proposed for learning disentangled representations, showing better performance than other unsupervised methods such as InfoGAN~\citep{chen2016infogan}. On the other hand, sVAE is trained using a similar variational objective as sRb-VAE, and can therefore be considered an unsupervised version of our method. We also evaluate vanilla VAEs \citep{kingma2013auto}. 

As discussed in Sec. \ref{sec:related}, there are no existing approaches in the literature that  directly  address reference-based disentangling. In order to evaluate an alternative weakly-supervised baseline exploiting the reference-set, we have implemented \citep{mathieu2016disentangling}, and adapted it to our context. Concretely, we have modified the learning algorithm in order to use only pairing information from reference imagesm by removing the reconstruction losses for  pairs of unlabelled samples as such  information is not available in reference-based disentangling. 

\input{Tables/table_all}

The different components of our method are implemented as deep neural networks. For this purpose, we have used conv-deconv architectures as is standard in VAE and GANs literature. Specifically, we employ the main building blocks used by \citet{karras2017progressive}, where the generator is implemented as a sequence of convolutions, Leaky-ReLU non-linearities, and nearest-neighbour up-sampling operations. Encoder and discriminators follow a similar architecture, using average pooling for down-sampling. See Appendix \ref{ap:architectures} for more details. For a fair comparison, we have developed our own implementation for all the evaluated methods in order to use the same network architectures and hyper-parameters. During optimization, we use the Adam optimizer~\citep{kingma2014adam} and a batch size of 36 images. For the MNIST and AffectNet , the models are learned for 30 and 20 epochs respectively. The number of latent variables for the encoders has been set to $32$ for all the experiments and models. The $\lambda$ parameter in the Laplace distribution is set to $0.01$. 

\subsection{Quantitative evaluation: Feature Learning}
\label{sec:quantitative}
A common strategy to evaluate the quality of learned representations is to measure the amount of information that they convey about the underlying generative factors. In our setting, we are interested in modelling the target factors that are constant  in the reference distribution.

\textbf{Experimental Setup.} Following a similar evaluation as \citet{mathieu2016disentangling}, we use the learned representations as feature vectors and train a low-capacity model estimating the target factors involved in each problem. Concretely, in the MNIST dataset we employ a set of linear-regressors predicting the scale, width and color parameters for each digit. To predict the different expression classes in the AffectNet dataset, we use a linear classifier. For methods using the reference-set, we used the inferred latent variables $\mathbf{e}$ as features since they are expected to encode the information regarding the target factors. In unsupervised models we use all the latent variables. For evaluation, we split each dataset in three subsets.  The first  is used to learn each generative model. Then, the second is used for training the regressors or classifier. Finally, the third is used to  evaluate the predictions in terms of the mean absolute error and per-class accuracy for the MNIST and AffectNet datasets, respectively. In MNIST, the second and third subset (5k images each) have been randomly generated from the original MNIST test set using the procedure described in Sec.~\ref{sec:datasets}. For AffectNet, we randomly select 500 images for each of the seven expressions from the original dataset, yielding 3,500 images per fold.

It is worth mentioning that some recent works \cite{kumar2017variational,chen2018isolating} have proposed alternative criterias to evaluate disentanglement. However, the proposed metrics are specifically designed to measure how a single dimension of the learned representation corresponds to a single ground-truth label. Note, however, that the one-to-one mapping assumption is not appropriate for real scenarios where we want to model high-level generative factors. For instance, it is unrealistic to expect that a single dimension of the latent vector $\mathbf{e}$ can convey all the information about a complex label such as the facial expression.

\begin{figure*}[t]
\centering
\includegraphics[width=0.95\linewidth]{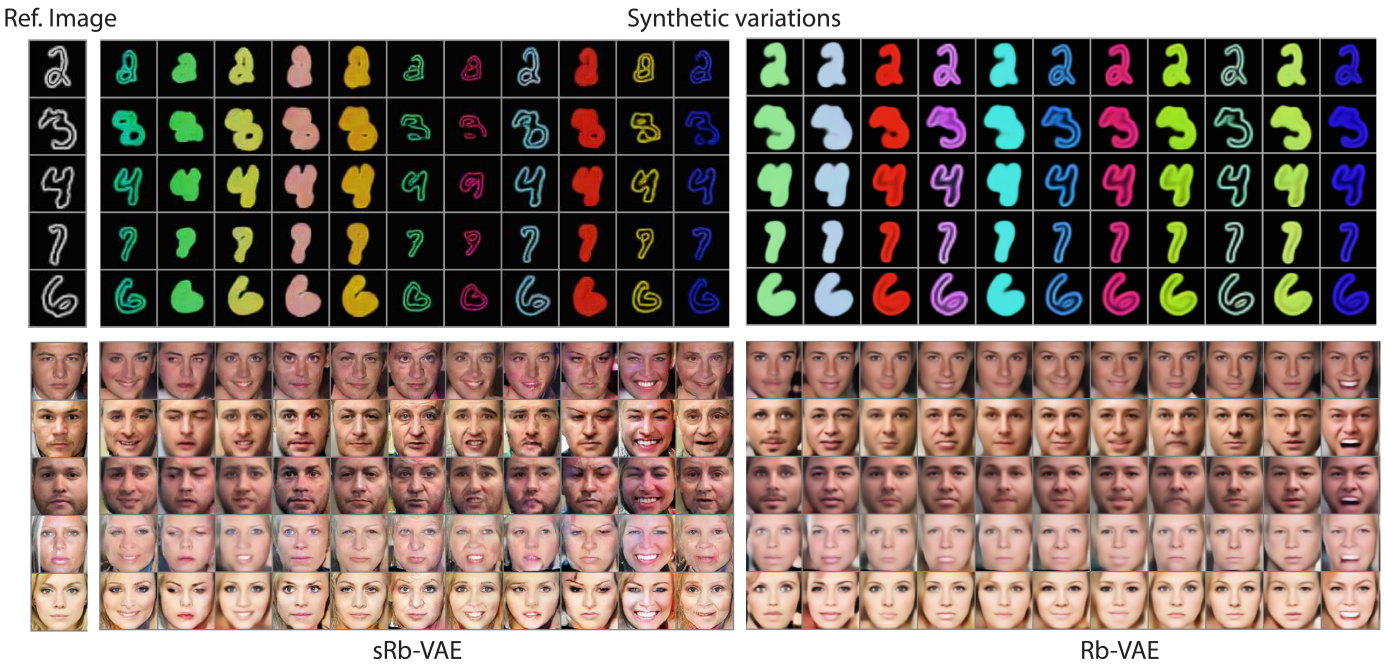}
\caption{Conditional image synthesis for MNIST (top) and AffectNet (bottom)  using sRb-VAE and Rb-VAE. Within each column images  are generated using the same random target factors $\mathbf{e}$.}
\label{fig:qual_cim}
\end{figure*}

\begin{figure}[h!]
\centering
\includegraphics[width=\linewidth]{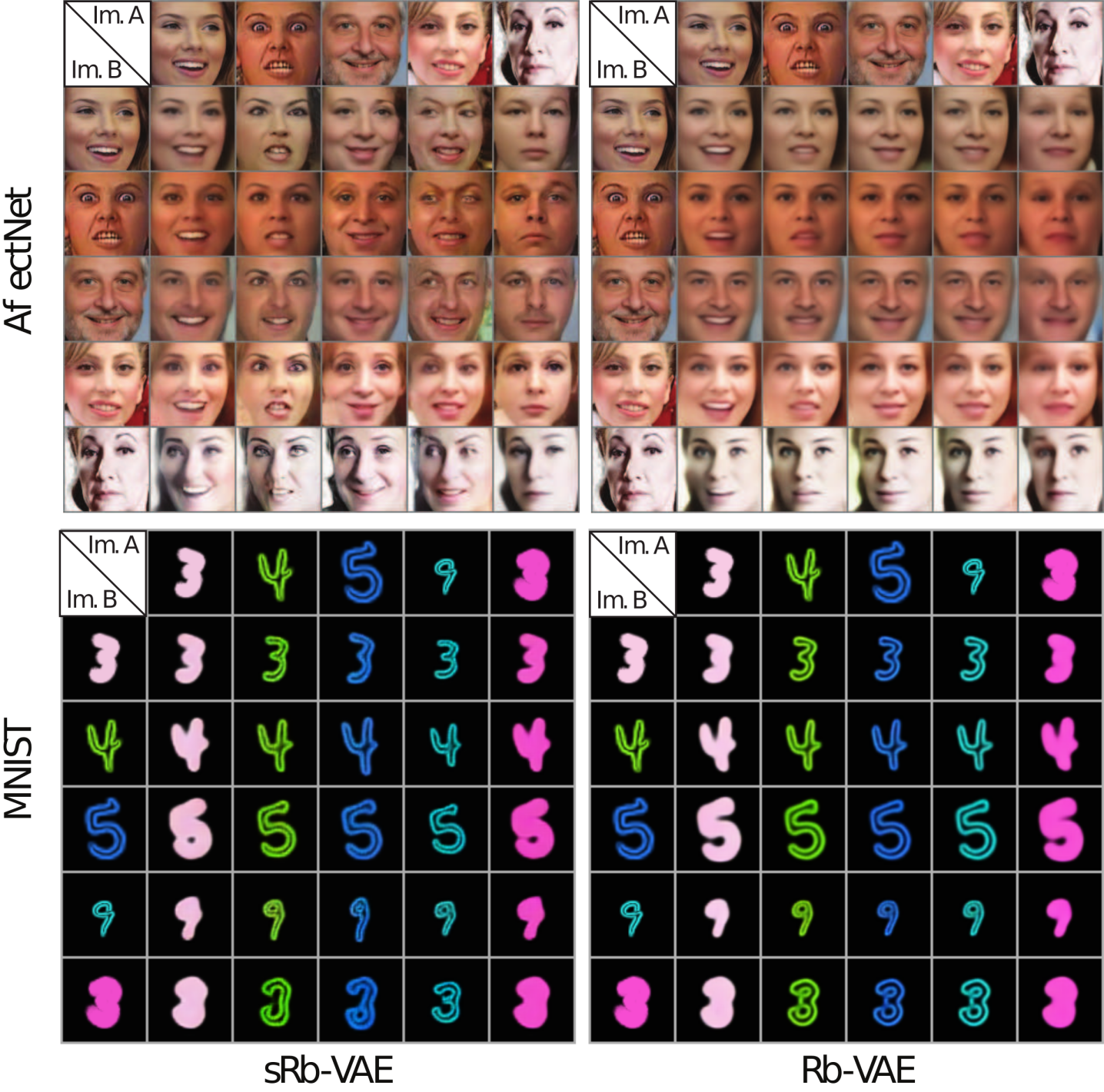}
\caption{Transferring target factors $\mathbf{e}$ from image A to an image B on AffectNet (expression) and MNIST (style).}
\label{fig:att_transfer}
\end{figure}

\textbf{Results and discussion.} Table \ref{tab:qual_results} shows the results obtained by the  different baselines considered and the proposed Rb-VAE and sRb-VAE. For DIP-VAE, $\beta$-VAE and $\beta$-TCVAE we tested different regularization parameters in the range $[1, 50]$, and report the best results. Note that the unsupervised approach DIP-VAE-I achieves better average results than Rb-VAE for MNIST. Moreover, in AffectNet, $\beta$-TCVAE achieves comparable or better performance in several cases. This may seem counter-intuitive because, unlike Rb-VAE, DIP-VAE-I is trained without the weak-supervision provided by reference images. However, it confirms our hypothesis that the learning objective of Rb-VAE does not explicitly encourage the disentanglement between target and common factors. In contrast, we can see that in most cases sRb-VAE obtains comparable or better results than rest of the methods. Moreover, it achieves the best average performance in both datasets. This demonstrates that the information provided by the reference distribution is effectively exploited by the symmetric KL objective used to train sRb-VAE. Additionally, note that the better performance of our model compared to unsupervised methods is informative. The reason is that the latter must encode all the generative factors into a single feature-vector. As a consequence, the target factors are entangled with the rest and the ground-truth labels are difficult to predict. In contrast, the representation $\mathbf{e}$ learned by our model is shown to be more effective because non-relevant factors are effectively removed, \ie encoded into $\mathbf{z}$.

In order to further validate this conclusion, we have followed the same evaluation protocol for Rb-VAE and sRb-VAE but considering the latent variables $\mathbf{z}$ as features. The average performance obtained by Rb-VAE is \textit{.349} and \textit{.195} for AffectNet and MNIST respectively. On the other hand, sRb-VAE achieves \textit{.335} and \textit{.189}.  Note that for both methods these results are significantly worse compared to using $\mathbf{e}$ as a representation in \tab{qual_results}. This shows that latent variable $\mathbf{z}$ is mainly modelling the common factors between reference and unlabelled images. The qualitative results presented in the next section confirm this. To conclude, note that sRb-VAE also obtains better performance than \citet{mathieu2016disentangling} in both data-sets. So even though this method also uses reference-images during training, sRb-VAE is shown to better exploit the weak-supervision existing in reference-based disentangling.

\subsection{Qualitative Evaluation}
\label{sec:qual_results}
In contrast to unsupervised methods, reference images are used by our model in order to split target and common factors into two different subsets of latent variables. This directly enables tasks such as conditional image synthesis or attribute transfer. In this section, we illustrate the potential applications of our proposed model in this settings.

\textbf{Conditional Image Synthesis.} The goal is to transform real images by modifying only the target factors $\mathbf{e}$. For instance, given a face of an individual, we aim to generate images of the same subject exhibiting different facial expressions. For this purpose, we use our model in order to infer the common factors $\mathbf{z}$. Then, we sample a vector $\mathbf{e}\sim \mathcal{N}(\mathbf{0},\mathbf{1})$ and use the generator network to obtain a new image from $\mathbf{e}$ and $\mathbf{z}$. In Fig.(\ref{fig:qual_cim}) we show examples of samples generated by Rb-VAE and sRb-VAE following this procedure. As we can observe, sRb-VAE generates more convincing results than its non-symmetric counterpart. In the AffectNet database, the amount of variability in Rb-VAE samples is quite low. In contrast, sRb-VAE is able to generate more diverse expressions related with eyes, mouth and eyebrows movements. Looking at the MNIST samples, we can draw similar conclusions. Whereas both methods generate transformations related with the digit color, Rb-VAE does not model scale variations in $\mathbf{e}$, while sRb-VAE does. This observation is coherent with  results reported in Tab.~\ref{tab:qual_results}, where Rb-VAE offers a poor estimation of the scale.

\textbf{Visual Attribute Transfer.} Here we transfer target factors $\mathbf{e}$ between a pair of images A and B. For example, given two samples from the MNIST dataset, the goal is to generate a new image with the digit in A modified with the style in B. Using our model, this can be easily achieved by  synthesizing a new image from latent variables $\mathbf{e}$ and $\mathbf{z}$ inferred from A and B respectively. Fig.~\ref{fig:att_transfer} shows images generated by sRb-VAE and Rb-VAE in this scenario.
In this case, we can draw similar conclusions than the previous experiment. Rb-VAE is not able to swap target factors related with the digit scale in the MNIST dataset, unlike sRb-VAE which better model this type of variation.  On the AffectNet images, both methods are able to keep most of the information regarding the identity of the subject, but again Rb-VAE leads to weaker expression changes than sRb-VAE. 

These qualitative results demonstrate that the standard variational objective of VAE is sub-optimal to train our model, and that the symmetric KL divergence objective used in sRb-VAE allows to better disentangle the common and target factors. Additional results are shown in Appendix \ref{ap:extra_results}.

%% file: Tables/table_all.tex
\begin{table*}[ht]
\centering
\resizebox{\textwidth}{!}
{
\begin{tabular}{c|ccccccc|c|ccccc|c|}
\multicolumn{1}{l|}{} & \multicolumn{8}{c|}{\textbf{AffectNet}} & \multicolumn{6}{c|}{\textbf{MNIST}} \\ \cline{2-15} 
\multicolumn{1}{l|}{\textbf{}} & \textbf{Happ} & \textbf{Sad} & \textbf{Sur} & \textbf{Fear} & \textbf{Disg} & \textbf{Ang} & \textbf{Compt} & \textbf{Avg.} & \textbf{R} & \textbf{G} & \textbf{B} & \textbf{Scale} & \textbf{Width} & \textbf{Avg.} \\ \hline
\rowcolor[HTML]{EFEFEF} 
\textbf{VAE} & .554 & .279 & .383 & .357 & .256 & .415 & .439 & .383 & .099 & .104 & .101 & \textbf{{[}.034{]}} & \textbf{.085} & .085 \\ \hline
\textbf{DIP-VAE-I} & .561 & .269 & \textbf{{[}.401{]}} & \textbf{.367} & .258 & .397 & .463 & .388 & \textbf{{[}.055{]}} & \textbf{.064} & .063 & .038 & .100 &  \textbf{.064} \\ \hline
\rowcolor[HTML]{EFEFEF} 
\textbf{DIP-VAE-II} & .548 & .245 & \textbf{{[}.401{]}} & \textbf{{[}.389{]}} & .268 & .391 & .463 & .386 & .077 & .069 & .076 & \textbf{.035} & .098 & .071 \\ \hline
\textbf{$\beta$ VAE} & .581 & .283 & .373 & .323 & .250 & .415 & .467 & .384 & .093 & .099 & .094 & .039 & .089 & .083 \\ \hline

\rowcolor[HTML]{EFEFEF} 
\textbf{sVAE} & \textbf{.583} & .251 & .389 & .349 & .260 & .391 & .469 & .384 & .094 & .092 & .084 & .036 & .104 & .082 \\ \hline

\rowcolor[HTML]{FFFFFF} 
\textbf{$\beta$-TCVAE} & .563 & .277 & \textbf{.393} & .349 & .256 & \textbf{{[}.427{]}} & .467 & .390 & .098 & .100 & .099 & \textbf{[.034]} & \textbf{[.084]} & .083 \\ \hline

\rowcolor[HTML]{EFEFEF} 
\textbf{[Mathieu et. al]} & .567 & .388 & .312 & .330 & .295 & .353 & \textbf{[.512]} & \textbf{.395} & .116 & .116 & .114 & .039 & .104 & .098 \\ \hline

\textbf{RBD-VAE} & .536 & \textbf{.393} & .379 & .311 & \textbf{.320} & .383 & .421 & 392 & .065 & .069 & \textbf{.062} & .061 & .095 & .070 \\ \hline

\rowcolor[HTML]{EFEFEF} 
{\color[HTML]{000000} \textbf{sRBD-VAE}} & {\color[HTML]{000000} \textbf{{[}.587{]}}} & {\color[HTML]{000000} \textbf{{[}.405{]}}} & {\color[HTML]{000000} .387} & {\color[HTML]{000000} .327} & {\color[HTML]{000000} \textbf{{[}.344{]}}} & {\color[HTML]{000000} \textbf{.425}} & {\color[HTML]{000000} \textbf{.483}} & {\color[HTML]{000000} \textbf{[.422]}} & {\color[HTML]{000000} \textbf{.057}} & {\color[HTML]{000000} \textbf{{[}.053{]}}} & {\color[HTML]{000000} \textbf{{[}.055{]}}} & {\color[HTML]{000000} .038} & {\color[HTML]{000000} .095} & {\color[HTML]{000000} \textbf{[.060]}}
\end{tabular}
}
\caption{Prediction of target factors from learned representations. We report accuracy and mean-absolute-error as evaluation metrics for the AffectNet and MNIST datasets, respectively. Two best methods shown in bold, best result in brackets. }
\label{tab:qual_results}
\end{table*}

%% file: supplementary.tex
\onecolumn
\section{Appendix}
\subsection{Mathematical derivations}
\label{ap:maths}

\textbf{Equivalence between $\mathbb{KL}\infdiv{q_\psi(\mathbf{x},\mathbf{z},\mathbf{e},y)}{p_\theta(\mathbf{x},\mathbf{z},\mathbf{e},y)}$ and Eq. (\textcolor{red}{3}):}

\begin{align}
& \sum_{y\in[0,1]} \int_{x,e,z} q_\psi(\mathbf{e} | \mathbf{x},y) q_\psi(\mathbf{z} | \mathbf{x}) p(\mathbf{x}|y) p(y) \log\Bigg( \frac{q_\psi(\mathbf{z},\mathbf{e} |\mathbf{x},y) p(\mathbf{x}|y) p(y)}{p_\theta(\mathbf{x}|\mathbf{e},\mathbf{z}) p(\mathbf{z}) p(\mathbf{e}|y)p(y)}   \Bigg) d\mathbf{x} d\mathbf{z} d\mathbf{e}  \\
& = \frac{1}{2} \int_{x,e,z} q_\psi(\mathbf{e} | \mathbf{x}) q_\psi(\mathbf{z} | \mathbf{x}) p^u(\mathbf{x}) \log\Bigg( \frac{q_\psi(\mathbf{e} | \mathbf{x}) q_\psi(\mathbf{z} | \mathbf{x}) p^u(\mathbf{x})}{p_\theta(\mathbf{x}|\mathbf{e},\mathbf{z}) p(\mathbf{z}) p(\mathbf{e})}   \Bigg) d\mathbf{x} d\mathbf{z} d\mathbf{e} \nonumber \\
& + \frac{1}{2} \int_{x,z} q_\psi(\mathbf{z} | \mathbf{x}) p^r(\mathbf{x}) \log\Bigg( \frac{q_\psi(\mathbf{z} | \mathbf{x}) p^r(\mathbf{x})}{p_\theta(\mathbf{x}|\mathbf{e}^r,\mathbf{z}) p(\mathbf{z})}   \Bigg) d\mathbf{x} d\mathbf{z}  \\
& = \frac{1}{2} \mathbb{E}_{p^u(\mathbf{x})} \mathbb{E}_{q_\psi(\mathbf{e} | \mathbf{x}) q_\psi(\mathbf{z} | \mathbf{x})} \Bigg[\log\Bigg( \frac{q_\psi(\mathbf{e} | \mathbf{x}) q_\psi(\mathbf{z} | \mathbf{x})}{p(\mathbf{z}) p(\mathbf{e})} \Bigg)  - \log(p_\theta(\mathbf{x}|\mathbf{e},\mathbf{z}))\Bigg] - H^u(\mathbf{x}) \nonumber \\  
& + \frac{1}{2} \mathbb{E}_{p^r(\mathbf{x})} \mathbb{E}_{q_\psi(\mathbf{z} | \mathbf{x})} \Bigg[\log\Bigg( \frac{q_\psi(\mathbf{z} | \mathbf{x})}{p(\mathbf{z}) } \Bigg)  - \log(p_\theta(\mathbf{x}|\mathbf{e}^r,\mathbf{z}))\Bigg] - H^r(\mathbf{x})  \\ 
& =  \frac{1}{2} \mathbb{E}_{ p^u(\mathbf{x})} \Bigg[\mathbb{KL}\infdiv{q_\psi( \mathbf{z} | \mathbf{x}) q_\psi( \mathbf{e} | \mathbf{x})}{p(\mathbf{z}) p(\mathbf{e})} - \mathbb{E}_{ q_\psi( \mathbf{z} | \mathbf{x}) q_\psi( \mathbf{e} | \mathbf{x})}  \log(p_\theta(\mathbf{x} | \mathbf{z} , \mathbf{e})) \Bigg] \nonumber \\
& + \frac{1}{2} \mathbb{E}_{ p^r(\mathbf{x})} \Bigg[\mathbb{KL}\infdiv{q_\psi( \mathbf{z} | \mathbf{x})}{p(\mathbf{z})} - \mathbb{E}_{ q_\psi( \mathbf{z} | \mathbf{x})}  \log(p_\theta(\mathbf{x} | \mathbf{z} , \mathbf{e}^r))\Bigg]  -  H^r(\mathbf{x}) - H^u(\mathbf{x}) \nonumber
\end{align}
We use  $H^r(X)$ and $H^u(X)$ to denote  the entropy of the reference and unlabelled distributions $p^r(\mathbf{x})$ and $p^u(\mathbf{x})$ respectively. Note that they can be ignored during the minimization since are constant \wrt parameters $\theta$ and $\psi$. For the second equality, we have used the definitions $p(\mathbf{x}|y=0)=p^u(\mathbf{x})$, $p(\mathbf{x}|y=1)=p^r(\mathbf{x})$ and assumed $p(y=0)=p(y=1)=\frac{1}{2}$. Moreover, we have exploited the fact that $q_\psi(\mathbf{e}|\mathbf{x},y=1)$ and $p(\mathbf{e}|y=1)$ are defined as delta functions and, therefore, $\mathbb{E}_{p(\mathbf{e}|y=1)}log(\frac{p(\mathbf{e}|y=1)}{q_\psi(\mathbf{e}|y=1)})=0$. We denote $p(\mathbf{e}|y=0)=p(\mathbf{e})$ and $q_\psi(\mathbf{e}|\mathbf{x},y=0)=q_\psi(\mathbf{e}|\mathbf{x})$ for the sake of brevity.

\textbf{Equivalence between $\mathbb{KL}\infdiv{p_\theta(\mathbf{x},\mathbf{z},\mathbf{e},y)}{q_\psi(\mathbf{x},\mathbf{z},\mathbf{e},y)}$ and the expression in Eq. (\textcolor{red}{5})}

\begin{align}
& \sum_{y\in[0,1]} \int_{x,e,z} p_\theta(\mathbf{x}|\mathbf{e},\mathbf{z}) p(\mathbf{z}) p(\mathbf{e}|y)p(y) \log\Bigg( \frac{p_\theta(\mathbf{x}|\mathbf{e},\mathbf{z}) p(\mathbf{z}) p(\mathbf{e}|y)p(y)}{q_\psi(\mathbf{z},\mathbf{e}|\mathbf{x},y) p(\mathbf{x}|y) p(y)}   \Bigg) d\mathbf{x} d\mathbf{z} d\mathbf{e}  \\
& = \frac{1}{2} \int_{x,e,z} p_\theta(\mathbf{x}|\mathbf{e},\mathbf{z}) p(\mathbf{z}) p(\mathbf{e}) \log\Bigg( \frac{p_\theta(\mathbf{x}|\mathbf{e},\mathbf{z}) p(\mathbf{z}) p(\mathbf{e})}{q_\psi(\mathbf{e} | \mathbf{x}) q_\psi(\mathbf{z} | \mathbf{x}) p^u(\mathbf{x})}   \Bigg) d\mathbf{x} d\mathbf{z} d\mathbf{e} \nonumber \\
& + \frac{1}{2} \int_{x,z} p_\theta(\mathbf{x}|\mathbf{e}^r,\mathbf{z}) p(\mathbf{z}) \log\Bigg( \frac{p_\theta(\mathbf{x}|\mathbf{e}^r,\mathbf{z}) p(\mathbf{z})}{q_\psi(\mathbf{z} | \mathbf{x}) p^r(\mathbf{x})}  \Bigg) d\mathbf{x} d\mathbf{z} d\mathbf{e} \\
& = \frac{1}{2} \mathbb{E}_{p(\mathbf{z}) p(\mathbf{e})} \mathbb{E}_{p_\theta(\mathbf{x}|\mathbf{e},\mathbf{z})} \Bigg[\log\Bigg( \frac{p_\theta(\mathbf{x}|\mathbf{e},\mathbf{z})}{p(\mathbf{x})^u} \Bigg)  - \log(q_\psi(\mathbf{e} | \mathbf{x}) q_\psi(\mathbf{z} | \mathbf{x}))\Bigg]  \nonumber \\ 
& + \frac{1}{2} \mathbb{E}_{p(\mathbf{z})} \mathbb{E}_{p_\theta(\mathbf{x}|\mathbf{e}^r,\mathbf{z})} \Bigg[\log\Bigg( \frac{p_\theta(\mathbf{x}|\mathbf{e}^r,\mathbf{z})}{p(\mathbf{x})^r} \Bigg)  - \log(q_\psi(\mathbf{z} | \mathbf{x}))\Bigg] - H(\mathbf{z}) - \frac{1}{2} H(\mathbf{e})\\  
& = \frac{1}{2} \mathbb{E}_{ p(\mathbf{z})p(\mathbf{e})} \Big[ \mathbb{KL}\infdiv{p_\theta(\mathbf{x} | \mathbf{z} , \mathbf{e})}{p^u(\mathbf{x})} - \mathbb{E}_{p_\theta(\mathbf{x} | \mathbf{z} , \mathbf{e})} [\log(q_\psi( \mathbf{z} | \mathbf{x})) + \log(q_\psi( \mathbf{e} | \mathbf{x}))]\Big] \nonumber \\
& +\frac{1}{2} \mathbb{E}_{ p(\mathbf{z})} \Big[\mathbb{KL}\infdiv{p_\theta(\mathbf{x} | \mathbf{z} , \mathbf{e}^r)}{p^r(\mathbf{x})} - \mathbb{E}_{p_\theta(\mathbf{x} | \mathbf{z} , \mathbf{e}^r)}  \log(q_\psi( \mathbf{z} | \mathbf{x}))\Big] - H(\mathbf{z}) - \frac{1}{2} H(\mathbf{e})
\end{align}

We have used the same definitions and assumptions previously discussed. Moreover, we
denote $H(\mathbf{z})$ and $H(\mathbf{e})$ as the entropy of the priors $p(\mathbf{z})$ and $p(\mathbf{e})$. Again, we can ignore these terms when we are optimizing w.r.t parameters $\psi$ and $\theta$.

\textbf{Equivalence between the minimization of the symmetric KL divergence in Eq. (\textcolor{red}{4}) and the expression in Eq. (\textcolor{red}{6})}

\begin{align}
& \mathbb{KL}\infdiv{q_\psi(\mathbf{z},\mathbf{e},\mathbf{x},y)}{p_\theta(\mathbf{x},\mathbf{z},\mathbf{e},y)} +  \mathbb{KL}\infdiv{p_\theta(\mathbf{x},\mathbf{z},\mathbf{e},y)}{q_\psi(\mathbf{z},\mathbf{e},\mathbf{x},y)} = \\
& = \mathbb{E}_{q_\psi(\mathbf{e} | \mathbf{x},y) q_\psi(\mathbf{z} | \mathbf{x}) p(\mathbf{x}|y) p(y) } \log\Bigg(\frac{q_\psi(\mathbf{e} | \mathbf{x},y) q_\psi(\mathbf{z} | \mathbf{x}) p(\mathbf{x}|y) p(y)}{p_\theta(\mathbf{x}|\mathbf{e},\mathbf{z}) p(\mathbf{z}) p(\mathbf{e}|y)p(y) }\Bigg) \nonumber \\
& + \mathbb{E}_{p_\theta(\mathbf{x}|\mathbf{e},\mathbf{z}) p(\mathbf{z}) p(\mathbf{e}|y)p(y) } \log\Bigg(\frac{p_\theta(\mathbf{x}|\mathbf{e},\mathbf{z}) p(\mathbf{z}) p(\mathbf{e}|y)p(y) }{q_\psi(\mathbf{e} | \mathbf{x},y) q_\psi(\mathbf{z} | \mathbf{x}) p(\mathbf{x}|y) p(y)}\Bigg)  \\
& = \frac{1}{2} \Bigg[\mathbb{E}_{q_\psi(\mathbf{e} , \mathbf{z} | \mathbf{x}) p^u(\mathbf{x})} \log\Bigg(\frac{q_\psi(\mathbf{e} , \mathbf{z} | \mathbf{x}) p^u(\mathbf{x})}{p_\theta(\mathbf{x}|\mathbf{e},\mathbf{z}) p(\mathbf{z}) p(\mathbf{e}) }\Bigg) + \mathbb{E}_{q_\psi(\mathbf{z} | \mathbf{x}) p^r(\mathbf{x}) } \log\Bigg(\frac{q_\psi(\mathbf{z} | \mathbf{x}) p(\mathbf{x})^r}{p_\theta(\mathbf{x}|\mathbf{e}^r,\mathbf{z}) p(\mathbf{z}) }\Bigg) \nonumber \\
& + \mathbb{E}_{p_\theta(\mathbf{x}|\mathbf{e},\mathbf{z}) p(\mathbf{z}) p(\mathbf{e}) } \log\Bigg(\frac{p_\theta(\mathbf{x}|\mathbf{e},\mathbf{z}) p(\mathbf{z}) p(\mathbf{e}) }{q_\psi(\mathbf{e},\mathbf{z} | \mathbf{x}) p^u(\mathbf{x})}\Bigg) + \mathbb{E}_{p_\theta(\mathbf{x}|\mathbf{e}^r,\mathbf{z}) p(\mathbf{z})} \log\Bigg(\frac{p_\theta(\mathbf{x}|\mathbf{e}^r,\mathbf{z}) p(\mathbf{z}))}{q_\psi(\mathbf{z} | \mathbf{x}) p^r(\mathbf{x})}\Bigg) \Bigg] \\
& = \frac{1}{2} \Big[\mathbb{E}_{q_\psi(\mathbf{e} , \mathbf{z} | \mathbf{x}) p^u(\mathbf{x})} \mathcal{L}_{\mathbf{x}\mathbf{z}\mathbf{e}} + \mathbb{E}_{q_\psi(\mathbf{z} | \mathbf{x}) p^r(\mathbf{x}) } \mathcal{L}_{\mathbf{x}\mathbf{z}} -  \mathbb{E}_{p_\theta(\mathbf{x}|\mathbf{e},\mathbf{z}) p(\mathbf{z}) p(\mathbf{e}) } \mathcal{L}_{\mathbf{x}\mathbf{z}\mathbf{e}} - \mathbb{E}_{p_\theta(\mathbf{x}|\mathbf{e}^r,\mathbf{z}) p(\mathbf{z})} \mathcal{L}_{\mathbf{x}\mathbf{z}}\Big]
\end{align}

\subsection{Pseudo-code for adversarial learning procedure}
\label{ap:alg}
Algorithm \ref{alg:learning} shows pseudo-code for the adversarial learning algorithm described in Sec. \textcolor{red}{4.3} of the paper.
\input{Others/learning_algorithm}


\subsection{Datasets}
\label{ap:datasets}

\begin{figure}[ht]
    \centering
    \includegraphics[width=0.65\textwidth]{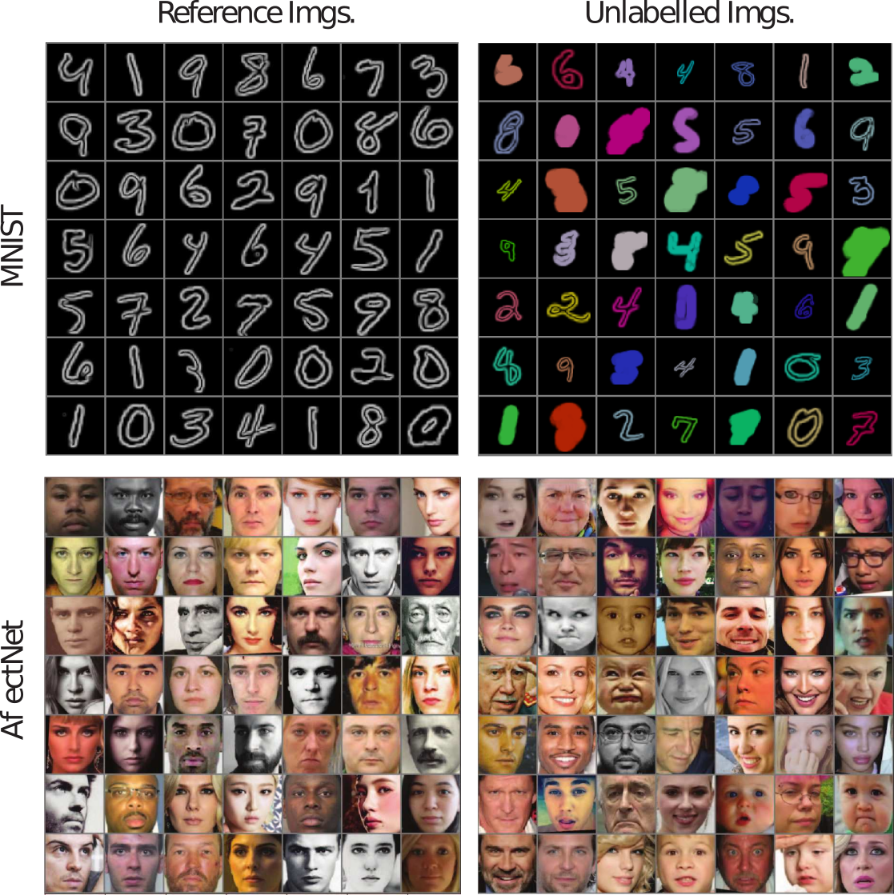}
    \caption{Examples of reference and unlabelled images used in our experiments. Extracted from MNIST (top) and AffectNet (bottom) datasets.}
    \label{fig:data}
\end{figure}

 Examples of reference and unlabelled images for MNIST and AffectNet are shown in Fig.~\ref{fig:data}.  
 In the  following, we provide more information about the used datasets.

\subsubsection{MNIST}

We use slightly modified version of the MNIST images: the size is increased to $64 \times 64$ pixels and an edge detection procedure is applied to keep only the boundaries of the digit.
We obtain the samples in the unlabelled dataset  by applying the following transformations over the MNIST images:
\begin{enumerate}
    \item \textbf{Width}: Generate a random integer in the range $\{1,\dots,10\}$ using a uniform distribution. Apply a dilation  operation over the image using a squared kernel with pixel-size equal to the generated number.
    \item \textbf{Color}: Generate a random 3D vector $c\in [0,1]^3$ using a uniform distribution. Normalize the resulting vector as $\hat{c}=c / ||c||_1$. Multiply the RGB components of all the pixels in the image by $\hat{c}$.
    \item \textbf{Size}: Generate a random number in the range $[0.5,1]$ using a uniform distribution. Downscale the image by a factor equal to the generated number. Apply zero-padding to the resulting image in order to recover the original resolution.
\end{enumerate}
 
 \subsubsection{AffectNet}
 
\textbf{Reference Set Collection.}
We collected a reference set of face images with neutral expression. We applied specific queries in order to obtain a large amount of faces from image search engines. Then, five different annotators filtered them in order to keep only images showing a neutral expression. The motivation for this data collection was that  we found that many neutral images in the AffectNet dataset~\cite{mollahosseini2017affectnet} are not accurate. As detailed in the original paper, the inter-observer agreement is significantly low for neutral images. In contrast, in our reference-set, each image was annotated in terms of ``neutral'' / ``non-neutral'' by two different annotators. In order to ensure a higher label quality compared to the AffectNet, only the images where both annotators agreed were added to the reference-set.

\textbf{Pre-processing.} In order to remove 2D affine transformations such as scaling or in-plane rotations, we apply an alignment process to the face images. We  localize facial landmarks using the approach of \citet{xiong2013supervised}. Then, we apply Procrustes analysis in order to find an affine transformation aligning the detected landmarks with a mean shape. Finally, we apply the transformation to the image and crop it. The resulting image is then re-sized to a resolution of $96 \times 96$ pixels. 

\subsection{Network architectures}
\label{ap:architectures}

\begin{figure}
\centering
\includegraphics[width=0.7\textwidth]{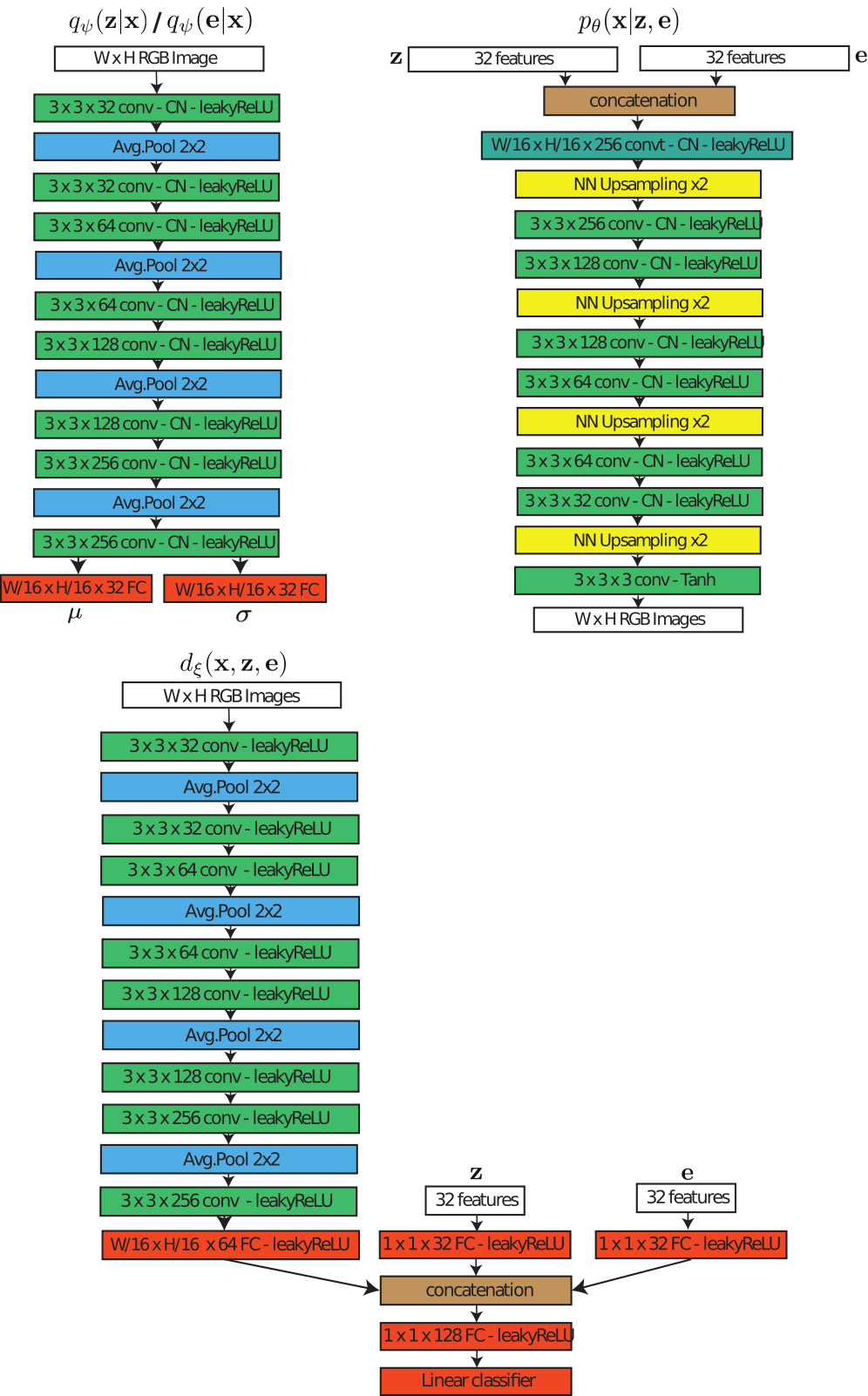}
\caption{Network architectures used in our experiments}
\label{fig:nw_arch}
\end{figure}

Fig. \ref{fig:nw_arch} illustrates the network architectures used in our experiments. CN refers to pixel-wise normalization as described in \citep{karras2017progressive}. FC defines a fully-connected layer. For Leaky ReLU non-linearities, we have used an slope of $0.2$. Given that we normalize the images in the range $[-1,1]$, we use an hyperbolic tangent function as the last layer of the generator. For the discriminator $d_\gamma(\mathbf{x},\mathbf{z})$, we use the same architecture showed for $d_\xi(\mathbf{x},\mathbf{z},\mathbf{e})$ but removing the input corresponding to $\mathbf{e}$. For the Adam optimizer~\citep{kingma2014adam} , we used $\alpha=10^{-4},\beta_1=0.5,\beta_2=0.99$ and $\epsilon=10^{-8}$. Note that the described architectures and hyper-parameters follow standard definitions according to most of GAN/VAEs previous works. 

In preliminary experiments, we found that the discriminator in sRb-VAE can start to ignore the inputs corresponding to latent variables $\mathbf{e}$  and $\mathbf{z}$ while focusing only on real and generated images. In order to mitigate this problem during training, we found it effective to randomly set to zero the inputs corresponding to latent variables and images of the last fully-connected layer. Note that this strategy is only used for sRB-VAE and sVAE in our experiments and it is not necessary in the other evaluated baselines. The reason is that these two methods are the only ones employing discriminators receiving images and features as input. We set the dropout probability to $0.25$. We found that this default value worked well for both methods in all the datasets and no specific fine-tuning of this hyper-parameter was necessary to mitigate the described phenomena.

\subsection{Additional Results}
\label{ap:extra_results}

\begin{figure}[ht]
    \centering
    \includegraphics[width=\textwidth]{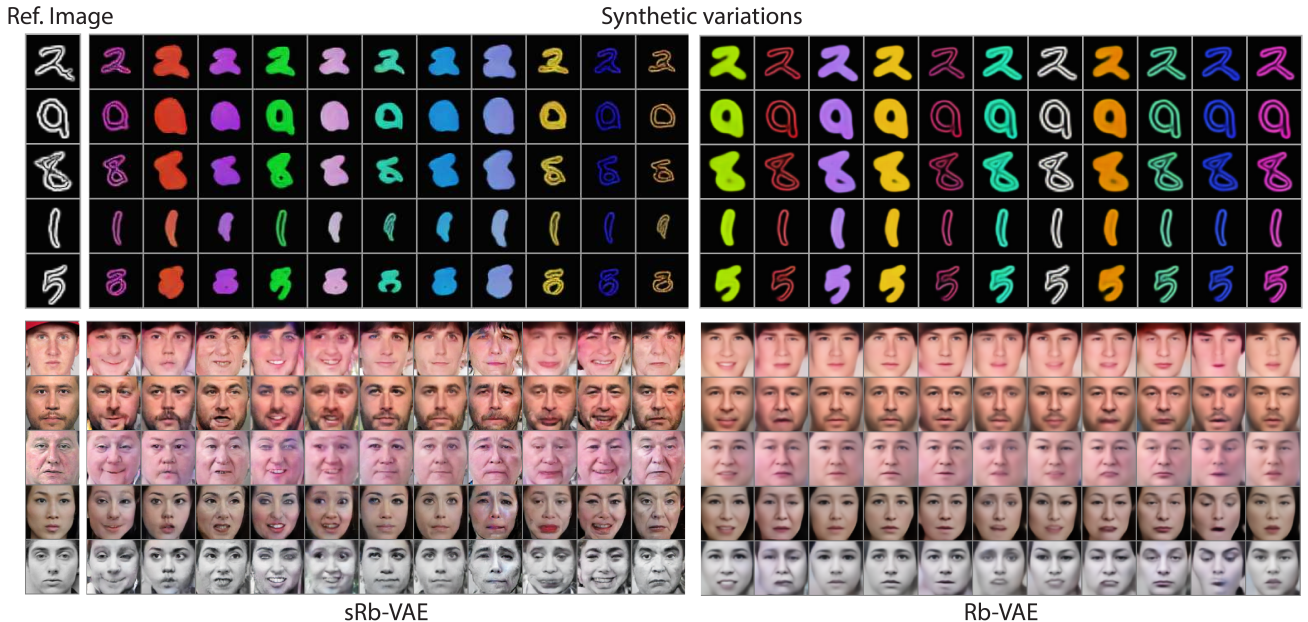}
    \caption{Qualitative results of sRb-VAE and Rb-VAE applied to conditional image generation. See Sec. (\textcolor{red}{5.4}) of the paper for details.}
    \label{fig:qual_supp_gen}
\end{figure}

\begin{figure}[h]
    \centering
    \includegraphics[width=0.65\textwidth]{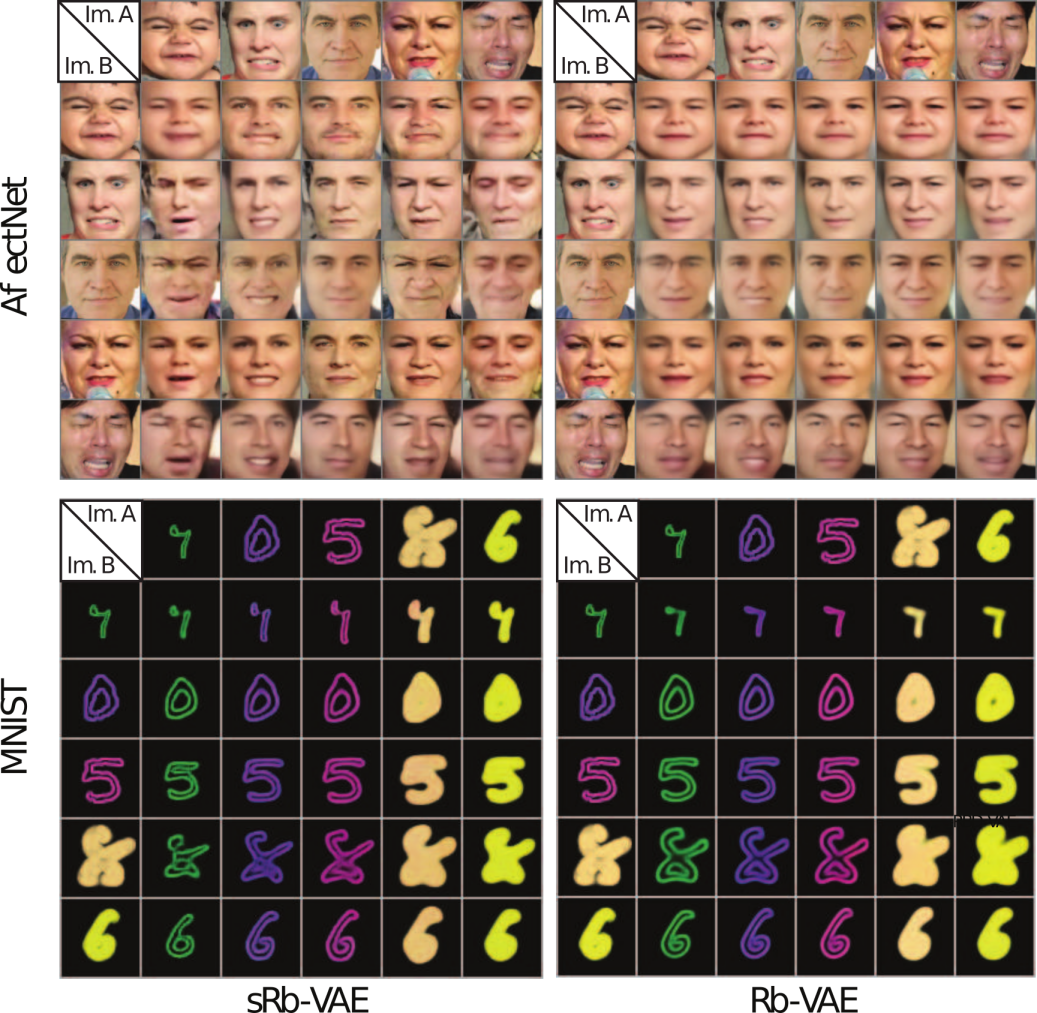}
    \caption{Qualitative results of sRb-VAE and Rb-VAE applied to visual attribute transfer. See Sec. (\textcolor{red}{5.4}) of the paper for details.}
    \label{fig:qual_supp_att_transf}
\end{figure}

\begin{figure}[ht]
    \centering
    \includegraphics[width=0.8\textwidth]{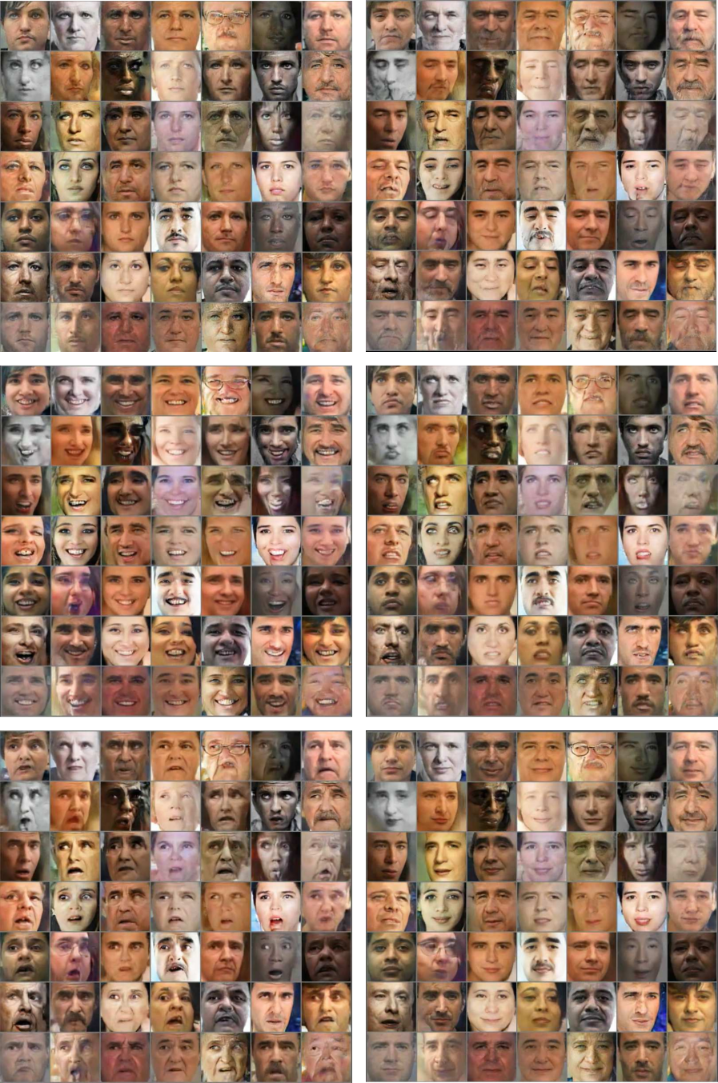}
    \caption{Images sample from the  sRB-VAE model. Images in the same panel share the same target factors $\mathbf{e}$ (expression). Images sharing the same position in the grids are generated from the same common factors $\mathbf{z}$ (identity) }
    \label{fig:qual_supp_gen_fx}
\end{figure}

Figures \ref{fig:qual_supp_gen} and \ref{fig:qual_supp_att_transf} show additional qualitative results for conditional image generation and visual attribute transfer, in the same spirit as  the figures in Section~\ref{sec:qual_results}. In order to provide more results for the conditional image generation task, we also provide two videos in this supplementary material. These videos contain animations generated by interpolating over the latent space corresponding to variations $\mathbf{e}$ (results shown for MNIST and AffectNet dataset). In Fig.~\ref{fig:qual_supp_gen_fx}, we also show additional images generated by sRB-VAE trained with the AffectNet dataset. Different from the previous cases, these images have been generated by just injecting random noise to the generator (over both latent variables $\mathbf{e}$ and $\mathbf{z}$). Note that different target factors $\mathbf{e}$ generate similar expressions in images generated from different common factors $\mathbf{z}$. The additional results further support the conclusions drawn in the main paper.

%% file: Others/learning_algorithm.tex
\begin{algorithm}[h]

\caption{sRb-VAE Advesarial Learning (Batch processing during SGD)}
\begin{multicols}{2}
\begin{algorithmic}[1]
{\scriptsize{

\STATE{\textit{/*** Gradient $\phi$  ***/}}
\STATE Sample $\{\mathbf{x}_1,...,\mathbf{x}_M\}$ from $p^u(\mathbf{x})$
\STATE Sample $\{\mathbf{x}^r_1,...,\mathbf{x}^r_M\}$ from $p^r(\mathbf{x})$
\STATE Sample $\{\mathbf{e}_1,...,\mathbf{e}_M\}$ using $q_\phi(\mathbf{e} | \mathbf{x})$
\STATE Sample $\{\mathbf{z}_1,...,\mathbf{z}_M\}$ using $q_\phi(\mathbf{z} | \mathbf{x})$
\STATE Sample $\{\mathbf{z}^r_1,...,\mathbf{z}^r_M\}$ using $q_\phi(\mathbf{z} | \mathbf{x}^r)$

\STATE Compute gradient of Eq. (\textcolor{red}{8}) w.r.t $\psi$ using the reparametrization trick for stochastic variables $\mathbf{z}$, $\mathbf{e}$ and $\mathbf{z}^r$: 
\begin{align}
g_\phi \leftarrow  \nabla_{\phi} \frac{1}{m} \Big[ & \sum_m d_\xi (\mathbf{{x}}_m, \mathbf{z}_m, \mathbf{e}_m) \nonumber \\ 
& + d_\gamma(\mathbf{{x}}^r_m, \mathbf{z}^r_m)\Big] \nonumber
\end{align}

\STATE{\textit{/*** Gradient $\theta$ ***/}}

\STATE Sample $\{\mathbf{\hat{e}}_1,...,\mathbf{\hat{e}}_M\}$ from $p(\mathbf{e})$
\STATE Sample $\{\mathbf{\hat{z}}_1,...,\mathbf{\hat{z}}_M\}$ from $p(\mathbf{z})$
\STATE Sample $\{\mathbf{\hat{z}}^r_1,...,\mathbf{\hat{z}}^r_M\}$ from $p(\mathbf{z})$
\STATE Sample $\{\mathbf{\hat{x}}_1,...,\mathbf{\hat{x}}_M\}$ using $p_\theta(\mathbf{x} | \mathbf{\hat{z}},\mathbf{e})$
\STATE Sample $\{\mathbf{\hat{x}}^r_1,...,\mathbf{\hat{x}}^r_M\}$ using $p_\theta(\mathbf{x} | \mathbf{\hat{z}},\mathbf{e}^r)$

\STATE Compute gradient of Eq. (\textcolor{red}{8}) w.r.t $\theta$ using the reparametrization trick for stochastic variables $\mathbf{\hat{x}}$ and $\mathbf{\hat{x}}^r$:

\begin{align}
g_\theta \leftarrow  \nabla_{\theta} \frac{1}{m} \Big[ & \sum_m d_\xi (\mathbf{\hat{x}}_m, \mathbf{\hat{z}}_m, \mathbf{\hat{e}}_m) \nonumber \\ 
& + d_\gamma(\mathbf{\hat{x}}^r_m, \mathbf{\hat{z}}^r_m)\Big] \nonumber
\end{align}

\vfill\null
\columnbreak

\STATE{\textit{/*** Gradient $\xi$ ***/}}
\STATE Compute gradient of discriminator function (Eq. (\textcolor{red}{9})) w.r.t $\xi$:
\begin{align}
g_\xi \leftarrow \nabla_{\xi} \frac{1}{2m} \sum_m \Big[ & \log(\sigma(d_\xi (\mathbf{{x}}_m, \mathbf{z}_m, \mathbf{e}_m)) +  \nonumber \\
 & \log(1-\sigma(d_\xi (\mathbf{\hat{x}}_m, \mathbf{\hat{z}}_m, \mathbf{\hat{e}}_m)) \Big] \nonumber
 \end{align}
 
\STATE{\textit{/*** Gradient $\gamma$ ***/}}
\STATE Compute gradient of discriminator function (Eq. (\textcolor{red}{9})) w.r.t $\gamma$:
\begin{align}
g_\gamma \leftarrow \nabla_{\gamma} \frac{1}{2m} \sum_m \Big[ & \log(\sigma(d_\gamma (\mathbf{{x}}^r_m, \mathbf{z}^r_m)) +  \nonumber \\
 & \log(1-\sigma(d_\gamma (\mathbf{\hat{x}}^r_m, \mathbf{\hat{z}}^r_m) \Big] \nonumber
 \end{align}
 
\STATE{\textit{/*** Update Parameters ***/}}
\STATE Update parameters via SGD with learning rate $\lambda$:
\begin{align}
&\theta \leftarrow \theta + \lambda g_\theta \nonumber \\
&\psi \leftarrow \psi - \lambda g_\psi \nonumber \\
&\xi \leftarrow \xi + \lambda g_\xi \nonumber \\
&\gamma \leftarrow \gamma + \lambda g_\gamma \nonumber
\end{align}

}
}
\end{algorithmic}
\end{multicols}
\label{alg:learning}

\end{algorithm}